\documentclass[10pt,twocolumn,letterpaper]{article}

\usepackage{iccv,times,graphicx,amsmath,amssymb,booktabs,tabulary,multirow,overpic,xcolor}
\usepackage[caption=false]{subfig}
\definecolor{citecolor}{RGB}{34,139,34}
\usepackage[pagebackref=true,breaklinks=true,letterpaper=true,colorlinks,
  citecolor=citecolor,bookmarks=false]{hyperref}
\def\iccvPaperID{***}\iccvfinalcopy

\usepackage[british,english,american]{babel}

\newcommand{\bd}[1]{\textbf{#1}}
\newcommand{\app}{\raise.17ex\hbox{$\scriptstyle\sim$}}
\newcommand{\ncdot}{{\mkern 0mu\cdot\mkern 0mu}}
\def\x{\times}
\newcolumntype{x}[1]{>{\centering\arraybackslash}p{#1pt}}
\newcommand{\dt}[1]{\fontsize{8pt}{.1em}\selectfont \emph{#1}}
\newlength\savewidth\newcommand\shline{\noalign{\global\savewidth\arrayrulewidth
  \global\arrayrulewidth 1pt}\hline\noalign{\global\arrayrulewidth\savewidth}}
\newcommand{\tablestyle}[2]{\setlength{\tabcolsep}{#1}\renewcommand{\arraystretch}{#2}\centering\footnotesize}
\makeatletter\renewcommand\paragraph{\@startsection{paragraph}{4}{\z@}
  {.5em \@plus1ex \@minus.2ex}{-.5em}{\normalfont\normalsize\bfseries}}\makeatother

\begin{document}

\title{Mask R-CNN}

\author{
 Kaiming He \quad Georgia Gkioxari \quad Piotr Doll\'ar \quad Ross Girshick \vspace{3mm}\\
 Facebook AI Research (FAIR) \vspace{-2mm}
}
\maketitle

\begin{abstract}
We present a conceptually simple, flexible, and general framework for object instance segmentation. Our approach efficiently detects objects in an image while simultaneously generating a high-quality segmentation mask for each instance. The method, called Mask R-CNN, extends Faster R-CNN by adding a branch for predicting an object mask in \emph{parallel} with the existing branch for bounding box recognition. Mask R-CNN is simple to train and adds only a small overhead to Faster R-CNN, running at 5 fps. Moreover, Mask R-CNN is easy to generalize to other tasks, \eg, allowing us to estimate human poses in the same framework. We show top results in all three tracks of the COCO suite of challenges, including instance segmentation, bounding-box object detection, and person keypoint detection. Without bells and whistles, Mask R-CNN outperforms all existing, single-model entries on every task, including the COCO 2016 challenge winners. We hope our simple and effective approach will serve as a solid baseline and help ease future research in instance-level recognition. Code has been made available at: \url{https://github.com/facebookresearch/Detectron}.
\end{abstract}

\section{Introduction}

The vision community has rapidly improved object detection and semantic segmentation results over a short period of time. In large part, these advances have been driven by powerful baseline systems, such as the Fast/Faster R-CNN \cite{Girshick2015a,Ren2015a} and Fully Convolutional Network (FCN) \cite{Long2015} frameworks for object detection and semantic segmentation, respectively. These methods are conceptually intuitive and offer flexibility and robustness, together with fast training and inference time. Our goal in this work is to develop a comparably enabling framework for \emph{instance segmentation}.

\begin{figure}[t]
\centering
\includegraphics[width=1\linewidth]{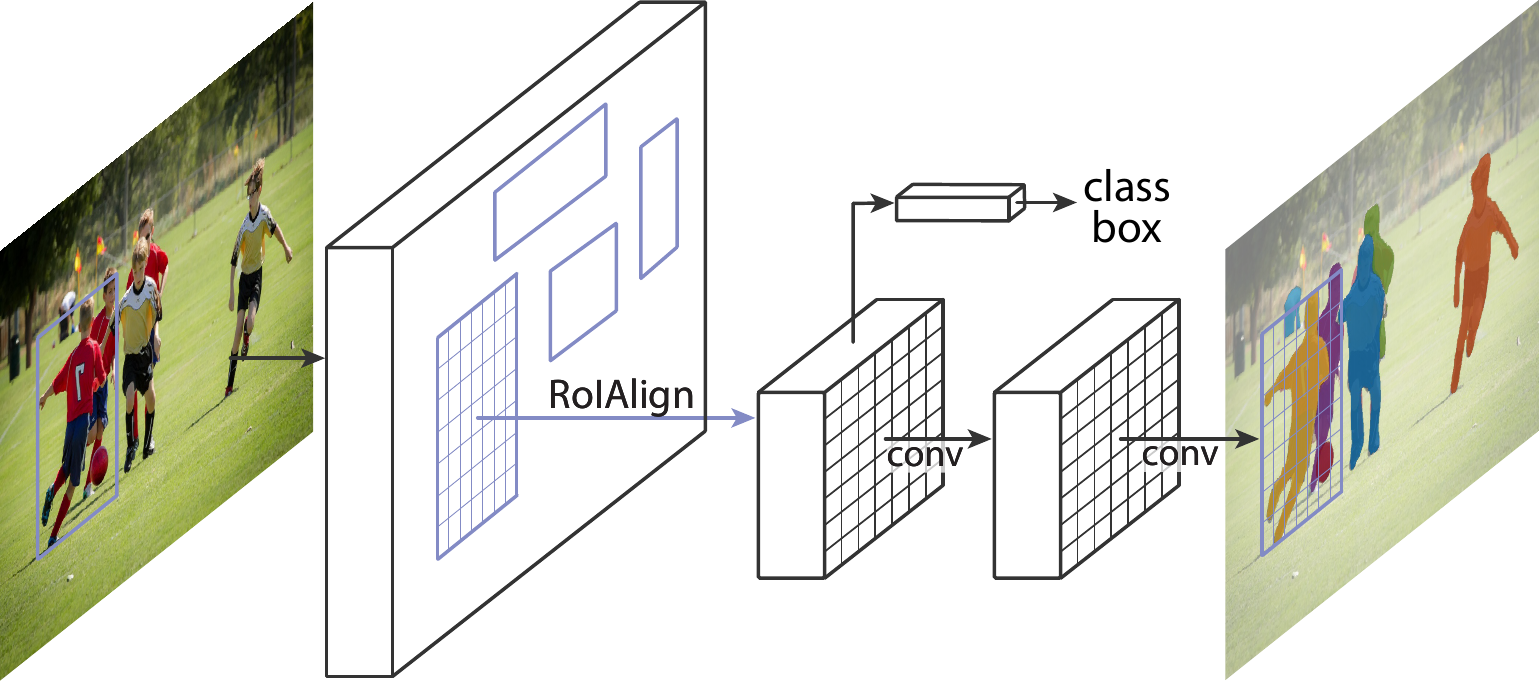}\vspace{2mm}
\caption{The \textbf{Mask\hspace{0.1297em}R-CNN} framework for instance segmentation.}
\label{fig:teaser}\vspace{-1mm}
\end{figure}

\begin{figure*}[t]
\centering
\includegraphics[width=1\linewidth]{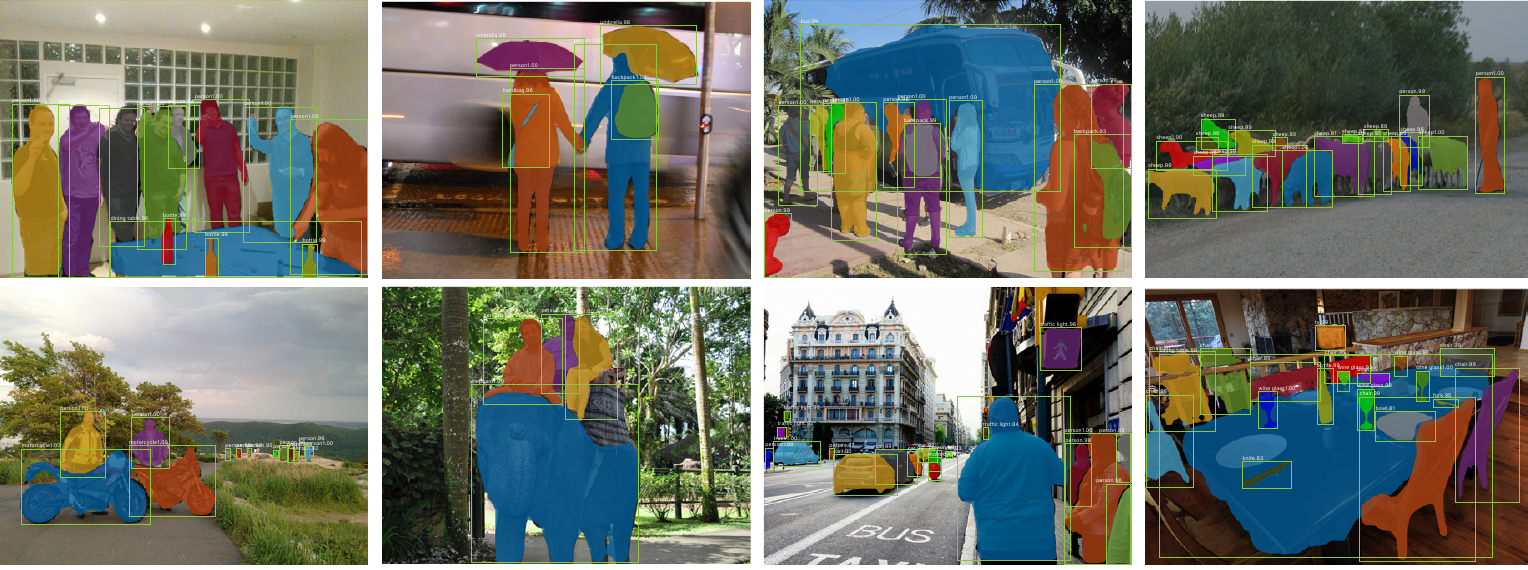}
\caption{\textbf{Mask R-CNN} results on the COCO test set. These results are based on ResNet-101 \cite{He2016}, achieving a \emph{mask} AP of 35.7 and running at 5 fps. Masks are shown in color, and bounding box, category, and confidences are also shown.}
\label{fig:results_main}\vspace{-2mm}
\end{figure*}

Instance segmentation is challenging because it requires the correct detection of all objects in an image while also precisely segmenting each instance. It therefore combines elements from the classical computer vision tasks of \emph{object detection}, where the goal is to classify individual objects and localize each using a bounding box, and \emph{semantic segmentation}, where the goal is to classify each pixel into a fixed set of categories without differentiating object instances.\footnote{Following common terminology, we use \emph{object detection} to denote detection via \emph{bounding boxes}, not masks, and \emph{semantic segmentation} to denote per-pixel classification without differentiating instances. Yet we note that \emph{instance segmentation} is both semantic and a form of detection.}~Given this, one might expect a complex method is required to achieve good results. However, we show that a surprisingly simple, flexible, and fast system can surpass prior state-of-the-art instance segmentation results.

Our method, called \emph{Mask R-CNN}, extends Faster R-CNN \cite{Ren2015a} by adding a branch for predicting segmentation masks on each Region of Interest (RoI), in \emph{parallel} with the existing branch for classification and bounding box regression (Figure~\ref{fig:teaser}). The mask branch is a small FCN applied to each RoI, predicting a segmentation mask in a pixel-to-pixel manner. Mask R-CNN is simple to implement and train given the Faster R-CNN framework, which facilitates a wide range of flexible architecture designs. Additionally, the mask branch only adds a small computational overhead, enabling a fast system and rapid experimentation.

In principle Mask R-CNN is an intuitive extension of Faster R-CNN, yet constructing the mask branch properly is critical for good results. Most importantly, Faster R-CNN was not designed for pixel-to-pixel alignment between network inputs and outputs. This is most evident in how \emph{RoIPool} \cite{He2014,Girshick2015a}, the \emph{de facto} core operation for attending to instances, performs coarse spatial quantization for feature extraction. To fix the misalignment, we propose a simple, quantization-free layer, called \emph{RoIAlign}, that faithfully preserves exact spatial locations. Despite being a seemingly minor change, RoIAlign has a large impact: it improves mask accuracy by relative 10\% to 50\%, showing bigger gains under stricter localization metrics. Second, we found it essential to \emph{decouple} mask and class prediction: we predict a binary mask for each class independently, without competition among classes, and rely on the network's RoI classification branch to predict the category. In contrast, FCNs usually perform per-pixel multi-class categorization, which couples segmentation and classification, and based on our experiments works poorly for instance segmentation.

Without bells and whistles, Mask R-CNN surpasses all previous state-of-the-art single-model results on the COCO instance segmentation task \cite{Lin2014}, including the heavily-engineered entries from the 2016 competition winner. As a by-product, our method also excels on the COCO object detection task. In ablation experiments, we evaluate multiple basic instantiations, which allows us to demonstrate its robustness and analyze the effects of core factors.

Our models can run at about 200ms per frame on a GPU, and training on COCO takes one to two days on a single 8-GPU machine. We believe the fast train and test speeds, together with the framework's flexibility and accuracy, will benefit and ease future research on instance segmentation.

Finally, we showcase the generality of our framework via the task of human pose estimation on the COCO keypoint dataset \cite{Lin2014}. By viewing each keypoint as a one-hot binary mask, with minimal modification Mask R-CNN can be applied to detect instance-specific poses. Mask R-CNN surpasses the winner of the 2016 COCO keypoint competition, and at the same time runs at 5 fps. Mask R-CNN, therefore, can be seen more broadly as a flexible framework for \emph{instance-level recognition} and can be readily extended to more complex tasks.

We have released code to facilitate future research.

\section{Related Work}

\paragraph{R-CNN:} The Region-based CNN (R-CNN) approach \cite{Girshick2014} to bounding-box object detection is to attend to a manageable number of candidate object regions \cite{Uijlings2013, Hosang2015} and evaluate convolutional networks  \cite{LeCun1989,Krizhevsky2012} independently on each RoI. R-CNN was extended \cite{He2014, Girshick2015a} to allow attending to RoIs on feature maps using RoIPool, leading to fast speed and better accuracy. Faster R-CNN \cite{Ren2015a} advanced this stream by learning the attention mechanism with a Region Proposal Network (RPN). Faster R-CNN is flexible and robust to many follow-up improvements (\eg, \cite{Shrivastava2016, Lin2017, Huang2017}), and is the current leading framework in several benchmarks.

\paragraph{Instance Segmentation:} Driven by the effectiveness of R-CNN, many approaches to instance segmentation are based on \emph{segment proposals}. Earlier methods \cite{Girshick2014, Hariharan2014, Hariharan2015, Dai2015} resorted to bottom-up segments \cite{Uijlings2013, Arbelaez2014}. DeepMask \cite{Pinheiro2015} and following works \cite{Pinheiro2016, Dai2016a} learn to propose segment candidates, which are then classified by Fast R-CNN. In these methods, segmentation \emph{precedes} recognition, which is slow and less accurate. Likewise, Dai \etal \cite{Dai2016} proposed a complex multiple-stage cascade that predicts segment proposals from bounding-box proposals, followed by classification. Instead, our method is based on \emph{parallel} prediction of masks and class labels, which is simpler and more flexible.

Most recently, Li \etal \cite{Li2017} combined the segment proposal system in \cite{Dai2016a} and object detection system in \cite{Dai2016b} for ``fully convolutional instance segmentation'' (FCIS). The common idea in \cite{Dai2016a, Dai2016b, Li2017} is to predict a set of position-sensitive output channels fully convolutionally. These channels simultaneously address object classes, boxes, and masks, making the system fast. But FCIS exhibits systematic errors on overlapping instances and creates spurious edges (Figure~\ref{fig:results_vs_fcis}), showing that it is challenged by the fundamental difficulties of segmenting instances.

Another family of solutions \cite{Kirillov2017,Bai2017,Arnab2017,Liu2017} to instance segmentation are driven by the success of semantic segmentation. Starting from per-pixel classification results (\eg, FCN outputs), these methods attempt to cut the pixels of the same category into different instances. In contrast to the \emph{segmentation-first} strategy of these methods, Mask R-CNN is based on an \emph{instance-first} strategy. We expect a deeper incorporation of both strategies will be studied in the future.

\section{Mask R-CNN}\label{sec:maskrcnn}

Mask R-CNN is conceptually simple: Faster R-CNN has two outputs for each candidate object, a class label and a bounding-box offset; to this we add a third branch that outputs the object mask. Mask R-CNN is thus a natural and intuitive idea. But the additional mask output is distinct from the class and box outputs, requiring extraction of much \emph{finer} spatial layout of an object. Next, we introduce the key elements of Mask R-CNN, including pixel-to-pixel alignment, which is the main missing piece of Fast/Faster R-CNN.

\paragraph{Faster R-CNN:} We begin by briefly reviewing the Faster R-CNN detector \cite{Ren2015a}. Faster R-CNN consists of two stages. The first stage, called a Region Proposal Network (RPN), proposes candidate object bounding boxes. The second stage, which is in essence Fast R-CNN \cite{Girshick2015a}, extracts features using RoIPool from each candidate box and performs classification and bounding-box regression. The features used by both stages can be shared for faster inference. We refer readers to \cite{Huang2017} for latest, comprehensive comparisons between Faster R-CNN and other frameworks.

\paragraph{Mask R-CNN:} Mask R-CNN adopts the same two-stage procedure, with an identical first stage (which is RPN). In the second stage, \emph{in parallel} to predicting the class and box offset, Mask R-CNN also outputs a binary mask for each RoI. This is in contrast to most recent systems, where classification \emph{depends} on mask predictions (\eg \cite{Pinheiro2015, Dai2016, Li2017}). Our approach follows the spirit of Fast R-CNN \cite{Girshick2015a} that applies bounding-box classification and regression in \emph{parallel} (which turned out to largely simplify the multi-stage pipeline of original R-CNN \cite{Girshick2014}).

Formally, during training, we define a multi-task loss on each sampled RoI as $L = L_{cls} + L_{box} + L_{mask}$. The classification loss $L_{cls}$ and bounding-box loss $L_{box}$ are identical as those defined in \cite{Girshick2015a}. The mask branch has a $Km^2$-dimensional output for each RoI, which encodes $K$ binary masks of resolution $m \x m$, one for each of the $K$ classes. To this we apply a per-pixel sigmoid, and define $L_{mask}$ as the average binary cross-entropy loss. For an RoI associated with ground-truth class $k$, $L_{mask}$ is only defined on the $k$-th mask (other mask outputs do not contribute to the loss).

Our definition of $L_{mask}$ allows the network to generate masks for every class without competition among classes; we rely on the dedicated classification branch to predict the class label used to select the output mask. This \emph{decouples} mask and class prediction. This is different from common practice when applying FCNs \cite{Long2015} to semantic segmentation, which typically uses a per-pixel \emph{softmax} and a \emph{multinomial} cross-entropy loss. In that case, masks across classes compete; in our case, with a per-pixel \emph{sigmoid} and a \emph{binary} loss, they do not. We show by experiments that this formulation is key for good instance segmentation results.

\paragraph{Mask Representation:} A mask encodes an input object's \emph{spatial} layout. Thus, unlike class labels or box offsets that are inevitably collapsed into short output vectors by fully-connected (\emph{fc}) layers, extracting the spatial structure of masks can be addressed naturally by the pixel-to-pixel correspondence provided by convolutions.

Specifically, we predict an $m \x m$ mask from each RoI using an FCN \cite{Long2015}. This allows each layer in the mask branch to maintain the explicit $m \x m$ object spatial layout without collapsing it into a vector representation that lacks spatial dimensions. Unlike previous methods that resort to \emph{fc} layers for mask prediction \cite{Pinheiro2015, Pinheiro2016, Dai2016}, our fully convolutional representation requires fewer parameters, and is more accurate as demonstrated by experiments.

This pixel-to-pixel behavior requires our RoI features, which themselves are small feature maps, to be well aligned to faithfully preserve the explicit per-pixel spatial correspondence. This motivated us to develop the following \emph{RoIAlign} layer that plays a key role in mask prediction.

\begin{figure}[t]
\begin{minipage}[c]{0.365\linewidth}
\includegraphics[width=\textwidth,trim={0 0 7.5mm 0},clip]{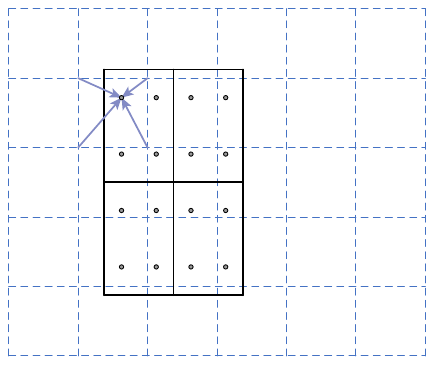}
\end{minipage}\hfill
\begin{minipage}[c]{0.605\linewidth}
\caption{\footnotesize \textbf{RoIAlign:} The dashed grid represents a feature map, the solid lines an RoI (with 2$\times$2 bins in this example), and the dots the 4 sampling points in each bin. RoIAlign computes the value of each sampling point by bilinear interpolation from the nearby grid points on the feature map. No quantization is performed on any coordinates involved in the RoI, its bins, or the sampling points.}
\label{fig:roialign}
\end{minipage}\vspace{-5mm}
\end{figure}

\paragraph{RoIAlign:} RoIPool \cite{Girshick2015a} is a standard operation for extracting a small feature map (\eg, 7$\x$7) from each RoI. RoIPool first \emph{quantizes} a floating-number RoI to the discrete granularity of the feature map, this quantized RoI is then subdivided into spatial bins which are themselves quantized, and finally feature values covered by each bin are aggregated (usually by max pooling). Quantization is performed, \eg, on a continuous coordinate $x$ by computing $[x/16]$, where 16 is a feature map stride and $[\cdot]$ is rounding; likewise, quantization is performed when dividing into bins (\eg, 7$\x$7). These quantizations introduce misalignments between the RoI and the extracted features. While this may not impact classification, which is robust to small translations, it has a large negative effect on predicting pixel-accurate masks.

To address this, we propose an \emph{RoIAlign} layer that removes the harsh quantization of RoIPool, properly \emph{aligning} the extracted features with the input. Our proposed change is simple: we avoid any quantization of the RoI boundaries or bins (\ie, we use $x/16$ instead of $[x/16]$). We use bilinear interpolation \cite{Jaderberg2015} to compute the exact values of the input features at four regularly sampled locations in each RoI bin, and aggregate the result (using max or average), see Figure~\ref{fig:roialign} for details. We note that the results are not sensitive to the exact sampling locations, or how many points are sampled, \emph{as long as} no quantization is performed.

RoIAlign leads to large improvements as we show in \S\ref{sec:ablations}. We also compare to the RoIWarp operation proposed in \cite{Dai2016}.
Unlike RoIAlign, RoIWarp overlooked the alignment issue and was implemented in \cite{Dai2016} as quantizing RoI just like RoIPool. So even though RoIWarp also adopts bilinear resampling motivated by \cite{Jaderberg2015}, it performs on par with RoIPool as shown by experiments (more details in Table~\ref{tab:ablation:roialign}), demonstrating the crucial role of alignment.

\paragraph{Network Architecture:} To demonstrate the generality of our approach, we instantiate Mask R-CNN with multiple architectures. For clarity, we differentiate between: (i) the convolutional \emph{backbone} architecture used for feature extraction over an entire image, and (ii) the network \emph{head} for bounding-box recognition (classification and regression) and mask prediction that is applied separately to each RoI.

We denote the \emph{backbone} architecture using the nomenclature \emph{network-depth-features}. We evaluate ResNet \cite{He2016} and ResNeXt \cite{Xie2017} networks of depth 50 or 101 layers. The original implementation of Faster R-CNN with ResNets \cite{He2016} extracted features from the final convolutional layer of the 4-th stage, which we call C4. This backbone with ResNet-50, for example, is denoted by ResNet-50-C4. This is a common choice used in \cite{He2016,Dai2016,Huang2017,Shrivastava2016a}.

We also explore another more effective backbone recently proposed by Lin \etal \cite{Lin2017}, called a Feature Pyramid Network (FPN). FPN uses a top-down architecture with lateral connections to build an in-network feature pyramid from a single-scale input. Faster R-CNN with an FPN backbone extracts RoI features from different levels of the feature pyramid according to their scale, but otherwise the rest of the approach is similar to vanilla ResNet. Using a ResNet-FPN backbone for feature extraction with Mask R-CNN gives excellent gains in both accuracy and speed. For further details on FPN, we refer readers to \cite{Lin2017}.

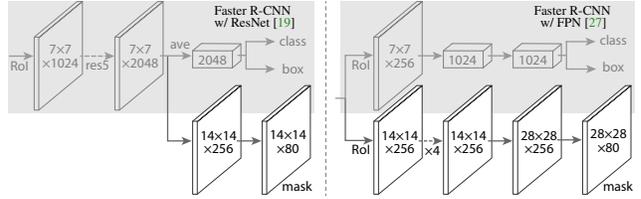
\begin{figure}[t]
\centering
\begin{overpic}[width=1.0\linewidth]{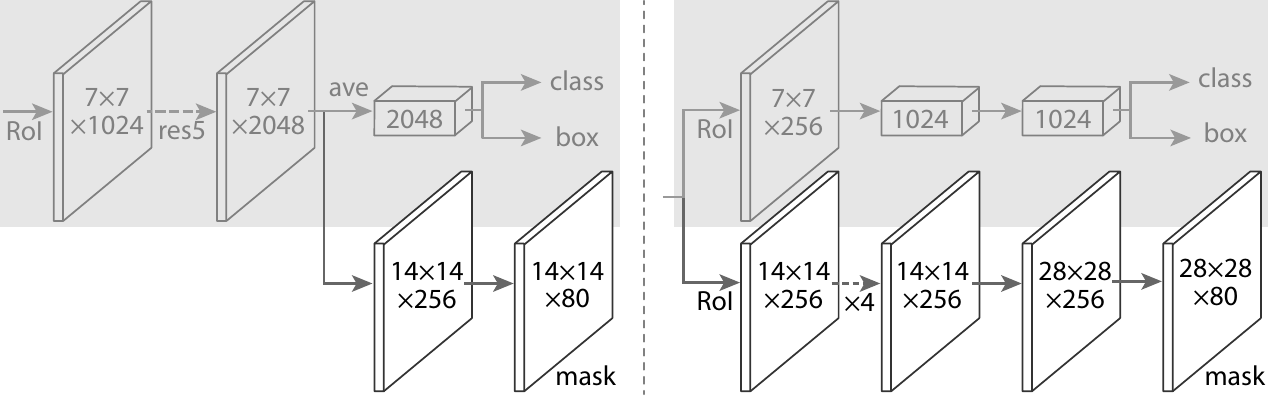}
 \put(33,29){\tiny Faster R-CNN}
 \put(33,27){\tiny w/ ResNet \cite{He2016}} 
 \put(84,29){\tiny Faster R-CNN}
 \put(85,27){\tiny w/ FPN \cite{Lin2017}}
\end{overpic}
\caption{\textbf{Head Architecture}: We extend two existing Faster R-CNN heads \cite{He2016,Lin2017}. Left/Right panels show the heads for the ResNet C4 and FPN backbones, from \cite{He2016} and \cite{Lin2017}, respectively, to which a mask branch is added.
Numbers denote spatial resolution and channels.
Arrows denote either conv, deconv, or \emph{fc} layers as can be inferred from context (conv preserves spatial dimension while deconv increases it).
All convs are 3$\x$3, except the output conv which is 1$\x$1, deconvs are 2$\x$2 with stride 2, and we use ReLU \cite{Nair2010} in hidden layers. \emph{Left}: `res5' denotes ResNet's fifth stage, which for simplicity we altered so that the first conv operates on a 7$\x$7 RoI with stride 1 (instead of 14$\x$14 / stride 2 as in \cite{He2016}). \emph{Right}: `$\x$4' denotes a stack of four consecutive convs.}
\label{fig:head}\vspace{-3mm}
\end{figure}

For the network \emph{head} we closely follow architectures presented in previous work to which we add a fully convolutional mask prediction branch. Specifically, we extend the Faster R-CNN box heads from the ResNet \cite{He2016} and FPN \cite{Lin2017} papers. Details are shown in Figure~\ref{fig:head}. The head on the ResNet-C4 backbone includes the 5-th stage of ResNet (namely, the 9-layer `res5' \cite{He2016}), which is compute-intensive. For FPN, the backbone already includes res5 and thus allows for a more efficient head that uses fewer filters.

We note that our mask branches have a straightforward structure. More complex designs have the potential to improve performance but are not the focus of this work.

\subsection{Implementation Details}\label{sec:impl}

We set hyper-parameters following existing Fast/Faster R-CNN work \cite{Girshick2015a, Ren2015a, Lin2017}. Although these decisions were made for object detection in original papers \cite{Girshick2015a, Ren2015a, Lin2017}, we found our instance segmentation system is robust to them.

\paragraph{Training:} As in Fast R-CNN, an RoI is considered positive if it has IoU with a ground-truth box of at least 0.5 and negative otherwise. The mask loss $L_{mask}$ is defined only on positive RoIs. The mask target is the intersection between an RoI and its associated ground-truth mask.

\begin{figure*}[t]
\centering
\includegraphics[width=1.0\linewidth]{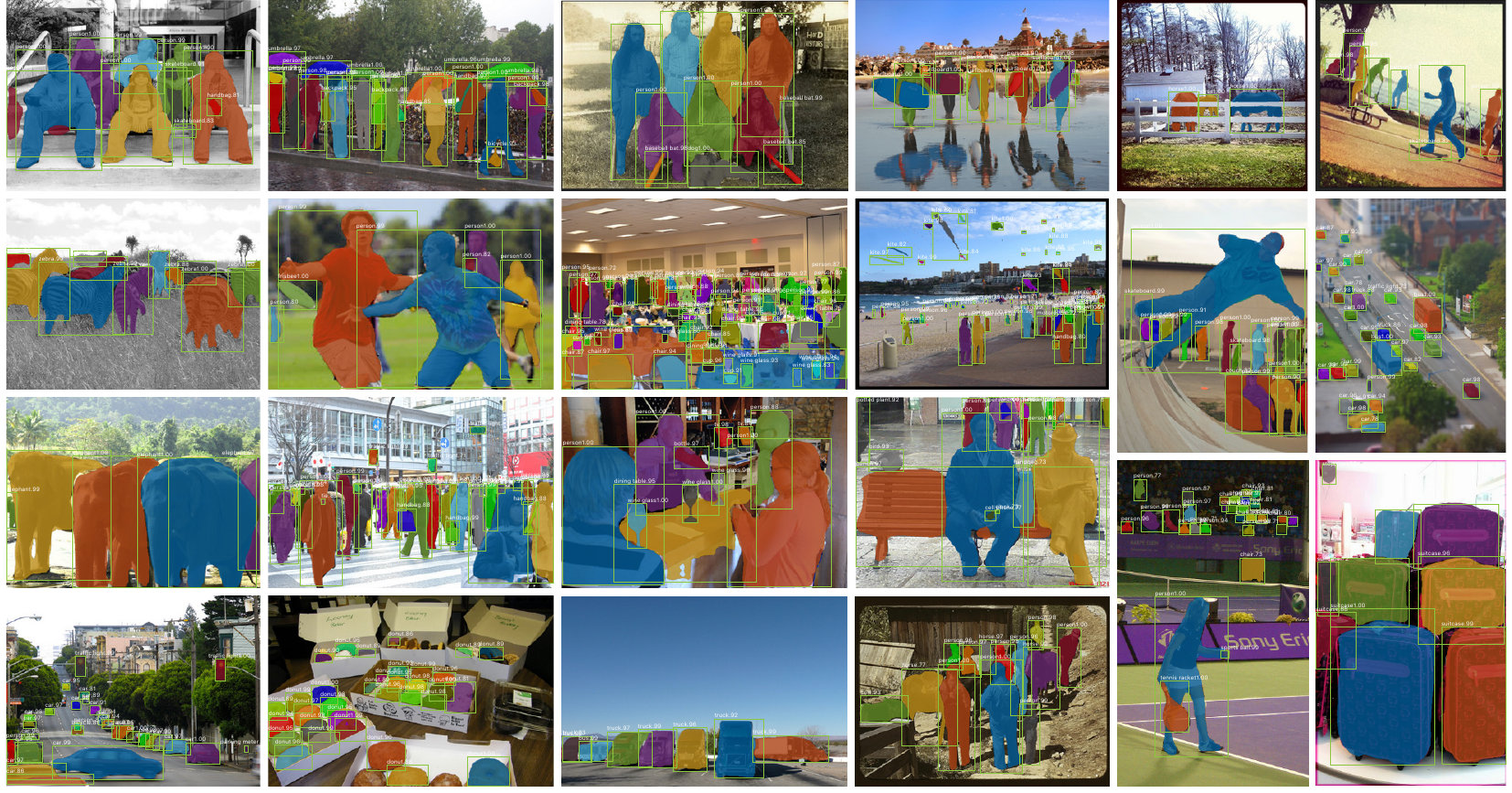}
\caption{More results of \textbf{Mask R-CNN} on COCO test images, using ResNet-101-FPN and running at 5 fps, with 35.7 mask AP (Table~\ref{tab:final_mask}).}
\label{fig:results_more}
\end{figure*}

\begin{table*}[t]
\tablestyle{3.5pt}{1.1}
\begin{tabular}{l|l|x{22}x{22}x{22}|x{22}x{22}x{22}}
 & backbone &  AP &  AP$_{50}$ & AP$_{75}$ & AP$_S$ &  AP$_M$ &  AP$_L$\\
\shline
 MNC \cite{Dai2016} & ResNet-101-C4
  & 24.6 & 44.3 & 24.8 & 4.7 & 25.9 & 43.6\\
 FCIS \cite{Li2017} +OHEM & ResNet-101-C5-dilated
  & 29.2 & 49.5 & - & 7.1 & 31.3 & 50.0\\
 FCIS+++ \cite{Li2017} +OHEM & ResNet-101-C5-dilated
  & 33.6 & 54.5 & - & - & - & -\\
\hline
 \bd{Mask R-CNN} & ResNet-101-C4
  & 33.1 & 54.9 & 34.8 & 12.1 & 35.6 & 51.1 \\
 \bd{Mask R-CNN} & ResNet-101-FPN
  & 35.7 & 58.0 & 37.8 & 15.5 & 38.1 & 52.4\\
 \bd{Mask R-CNN} & ResNeXt-101-FPN
  & \bd{37.1} & \bd{60.0} & \bd{39.4} & \bd{16.9} & \bd{39.9} & \bd{53.5}
\end{tabular}\vspace{2mm}
\caption{\textbf{Instance segmentation} \emph{mask} AP on COCO \texttt{test-dev}. MNC \cite{Dai2016} and FCIS \cite{Li2017} are the winners of the COCO 2015 and 2016 segmentation challenges, respectively. Without bells and whistles, Mask R-CNN outperforms the more complex FCIS+++, which includes multi-scale train/test, horizontal flip test, and OHEM \cite{Shrivastava2016}. All entries are \emph{single-model} results.}
\label{tab:final_mask}\vspace{-3mm}
\end{table*}

We adopt image-centric training \cite{Girshick2015a}. Images are resized such that their scale (shorter edge) is 800 pixels \cite{Lin2017}. Each mini-batch has 2 images per GPU and each image has $N$ sampled RoIs, with a ratio of 1:3 of positive to negatives \cite{Girshick2015a}. $N$ is 64 for the C4 backbone (as in \cite{Girshick2015a,Ren2015a}) and 512 for FPN (as in \cite{Lin2017}).  We train on 8 GPUs (so effective mini-batch size is 16) for 160k iterations, with a learning rate of 0.02 which is decreased by 10 at the 120k iteration. We use a weight decay of 0.0001 and momentum of 0.9. With ResNeXt \cite{Xie2017}, we train with 1 image per GPU and the same number of iterations, with a starting learning rate of 0.01.

The RPN anchors span 5 scales and 3 aspect ratios, following \cite{Lin2017}. For convenient ablation, RPN is trained separately and does not share features with Mask R-CNN, unless specified. For every entry in this paper, RPN and Mask R-CNN have the same backbones and so they are shareable.

\paragraph{Inference:} At test time, the proposal number is 300 for the C4 backbone (as in \cite{Ren2015a}) and 1000 for FPN (as in \cite{Lin2017}). We run the box prediction branch on these proposals, followed by non-maximum suppression \cite{Girshick2015}. The mask branch is then applied to the highest scoring 100 detection boxes. Although this differs from the parallel computation used in training, it speeds up inference and improves accuracy (due to the use of fewer, more accurate RoIs). The mask branch can predict $K$ masks per RoI, but we only use the $k$-th mask, where $k$ is the predicted class by the classification branch. The $m$$\times$$m$ floating-number mask output is then resized to the RoI size, and binarized at a threshold of 0.5.

Note that since we only compute masks on the top 100 detection boxes, Mask R-CNN adds a small overhead to its Faster R-CNN counterpart (\eg, \app20\% on typical models).

\section{Experiments: Instance Segmentation}\label{sec:results}

We perform a thorough comparison of Mask R-CNN to the state of the art along with comprehensive ablations on the COCO dataset \cite{Lin2014}. We report the standard COCO metrics including AP (averaged over IoU thresholds), AP$_{50}$, AP$_{75}$, and AP$_S$, AP$_M$, AP$_L$ (AP at different scales). Unless noted, AP is evaluating using \emph{mask} IoU. As in previous work \cite{Bell2016, Lin2017}, we train using the union of 80k train images and a 35k subset of val images (\texttt{trainval35k}), and report ablations on the remaining 5k val images (\texttt{minival}). We also report results on \texttt{test-dev} \cite{Lin2014}.

\subsection{Main Results}

We compare Mask R-CNN to the state-of-the-art methods in instance segmentation in Table \ref{tab:final_mask}. All instantiations of our model outperform baseline variants of previous state-of-the-art models. This includes MNC \cite{Dai2016} and FCIS \cite{Li2017}, the winners of the COCO 2015 and 2016 segmentation challenges, respectively. Without bells and whistles, Mask R-CNN with ResNet-101-FPN backbone outperforms FCIS+++ \cite{Li2017}, which includes multi-scale train/test, horizontal flip test, and online hard example mining (OHEM) \cite{Shrivastava2016}. While outside the scope of this work, we expect many such improvements to be applicable to ours.

\begin{figure*}[t]
\centering
\includegraphics[width=1.0\linewidth]{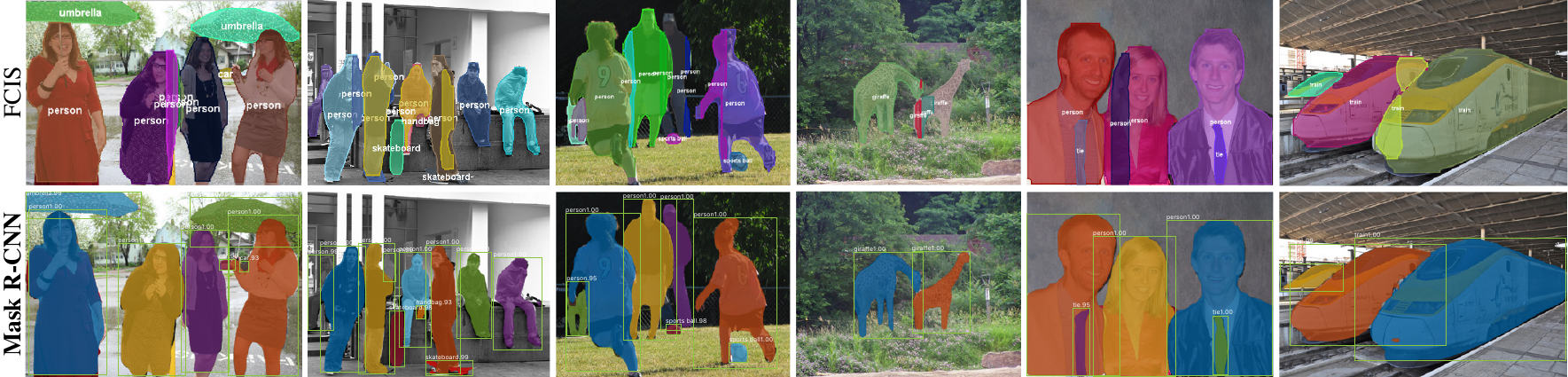}
\caption{FCIS+++ \cite{Li2017} (top) \vs Mask R-CNN (bottom, ResNet-101-FPN). FCIS exhibits systematic artifacts on overlapping objects.}
\label{fig:results_vs_fcis}
\end{figure*}

Mask R-CNN outputs are visualized in Figures \ref{fig:results_main} and \ref{fig:results_more}. Mask R-CNN achieves good results even under challenging conditions. In Figure \ref{fig:results_vs_fcis} we compare our Mask R-CNN baseline and FCIS+++ \cite{Li2017}. FCIS+++ exhibits systematic artifacts on overlapping instances, suggesting that it is challenged by the fundamental difficulty of instance segmentation. Mask R-CNN shows no such artifacts.

\begin{table*}[t]\vspace{-3mm}
\subfloat[\textbf{Backbone Architecture}: Better backbones bring expected gains: deeper networks do better, FPN outperforms C4 features, and ResNeXt improves on ResNet.\label{tab:ablation:backbone}]{
\tablestyle{2.5pt}{1.05}\begin{tabular}{c|x{22}x{22}x{22}}
 \scriptsize \emph{net-depth-features} & AP & AP$_{50}$ & AP$_{75}$\\
\shline
 \scriptsize ResNet-50-C4 & 30.3 & 51.2 & 31.5\\
 \scriptsize ResNet-101-C4 & 32.7 & 54.2 & 34.3\\\hline
 \scriptsize ResNet-50-FPN & 33.6 & 55.2 & 35.3\\
 \scriptsize ResNet-101-FPN & 35.4 & 57.3 & 37.5\\
 \scriptsize ResNeXt-101-FPN & \bd{36.7} & \bd{59.5} & \bd{38.9}
\end{tabular}}\hspace{3mm}
\subfloat[\textbf{Multinomial \vs Independent Masks} (ResNet-50-C4): \emph{Decoupling} via per-class binary masks (sigmoid) gives large gains over multinomial masks (softmax).\label{tab:ablation:sigmoid}]{
\tablestyle{4.8pt}{1.05}\begin{tabular}{c|x{22}x{22}x{22}}
 & AP & AP$_{50}$ & AP$_{75}$\\
\shline
 \emph{softmax} & 24.8 & 44.1 & 25.1\\
 \emph{sigmoid} & \bd{30.3} & \bd{51.2} & \bd{31.5}\\
\hline
 & \dt{+5.5} & \dt{+7.1} & \dt{+6.4}\\
 \multicolumn{4}{c}{~}\\
 \multicolumn{4}{c}{~}\\
\end{tabular}}\hspace{3mm}
\subfloat[\textbf{RoIAlign} (ResNet-50-C4): Mask results with various RoI layers. Our RoIAlign layer improves AP by $\app$3 points and AP$_{75}$ by $\app$5 points. Using proper alignment is the only factor that contributes to the large gap between RoI layers.\label{tab:ablation:roialign}]{
\tablestyle{2.2pt}{1.05}\begin{tabular}{c|c|c|c|x{22}x{22}x{22}}
 & \scriptsize\textbf{align?} & \scriptsize bilinear? & \scriptsize agg. 
 & AP & AP$_{50}$ & AP$_{75}$\\
\shline
 \emph{RoIPool} \cite{Girshick2015a}
  & & & max & 26.9 & 48.8 & 26.4\\
\hline
 \multirow{2}{*}{\emph{RoIWarp} \cite{Dai2016}}
  & & \checkmark & max & 27.2 & 49.2 & 27.1\\
  & & \checkmark & ave & 27.1 & 48.9 & 27.1\\
\hline
 \multirow{2}{*}{\emph{RoIAlign}}
  & \checkmark & \checkmark & max & \bd{30.2} & \bd{51.0} & \bd{31.8}\\
  & \checkmark & \checkmark & ave & \bd{30.3} & \bd{51.2} & \bd{31.5}
\end{tabular}}\vspace{-1mm}\\
\subfloat[\textbf{RoIAlign} (ResNet-50-\bd{C5}, \emph{stride 32}): Mask-level and box-level AP using \emph{large-stride} features. Misalignments are more severe than with stride-16 features (Table \ref{tab:ablation:roialign}), resulting in big accuracy gaps.\label{tab:ablation:roialign32}]{
\tablestyle{4pt}{1.05}\begin{tabular}{c|x{22}x{22}x{22}|x{22}x{22}x{22}}
 & AP & AP$_{50}$ & AP$_{75}$
 & AP$^\text{bb}$ & AP$^\text{bb}_{50}$ & AP$^\text{bb}_{75}$ \\[.1em]
\shline
 \emph{RoIPool} & 23.6 & 46.5 & 21.6 & 28.2 & 52.7 & 26.9\\
 \emph{RoIAlign} & \bd{30.9} & \bd{51.8} & \bd{32.1} & \bd{34.0} & \bd{55.3} & \bd{36.4}\\
\hline
 & \dt{+7.3} & \dt{+ 5.3} & \dt{+10.5} & \dt{+5.8} & \dt{+2.6} & \dt{+9.5}
\end{tabular}}\hspace{5mm}
\subfloat[\textbf{Mask Branch} (ResNet-50-FPN): Fully convolutional networks (FCN) \vs multi-layer perceptrons (MLP, fully-connected) for mask prediction. FCNs improve results as they take advantage of explicitly encoding spatial layout.\label{tab:ablation:maskhead}]{
\tablestyle{4pt}{1.05}\begin{tabular}{c|c|x{22}x{22}x{22}}
 & mask branch & AP & AP$_{50}$ & AP$_{75}$\\
\shline
 MLP & fc: 1024$\rightarrow$1024$\rightarrow$$80\ncdot28^2$  & 31.5 & 53.7 & 32.8\\
 MLP & fc: 1024$\rightarrow$1024$\rightarrow$1024$\rightarrow$$80\ncdot28^2$ & 31.5 & 54.0 & 32.6\\
\hline
 \textbf{FCN} &  conv: 256$\rightarrow$256$\rightarrow$256$\rightarrow$256$\rightarrow$256$\rightarrow$80 
 & \bd{33.6} & \bd{55.2} & \bd{35.3}
\end{tabular}}\vspace{2mm}
\caption{\textbf{Ablations}. We train on \texttt{trainval35k}, test on \texttt{minival}, and report \emph{mask} AP unless otherwise noted.}
\label{tab:ablations}\vspace{-3mm}
\end{table*}

\subsection{Ablation Experiments}\label{sec:ablations}

We run a number of ablations to analyze Mask R-CNN. Results are shown in Table \ref{tab:ablations} and discussed in detail next.

\paragraph{Architecture:} Table \ref{tab:ablation:backbone} shows Mask R-CNN with various backbones. It benefits from deeper networks (50 \vs~101) and advanced designs  including FPN and ResNeXt. We note that \emph{not} all frameworks automatically benefit from deeper or advanced networks (see benchmarking in \cite{Huang2017}).

\paragraph{Multinomial \vs Independent Masks:} Mask R-CNN \emph{decouples} mask and class prediction: as the existing box branch predicts the class label, we generate a mask for each class without competition among classes (by a per-pixel \emph{sigmoid} and a \emph{binary} loss). In Table \ref{tab:ablation:sigmoid}, we compare this to using a per-pixel \emph{softmax} and a \emph{multinomial} loss (as commonly used in FCN \cite{Long2015}). This alternative \emph{couples} the tasks of mask and class prediction, and results in a severe loss in mask AP (5.5 points). This suggests that once the instance has been classified as a whole (by the box branch), it is sufficient to predict a binary mask without concern for the categories, which makes the model easier to train.

\paragraph{Class-Specific \vs Class-Agnostic Masks:} Our default instantiation predicts class-specific masks, \ie, one $m$$\x$$m$ mask per class. Interestingly, Mask R-CNN with class-agnostic masks (\ie, predicting a single $m$$\x$$m$ output regardless of class) is nearly as effective: it has 29.7 mask AP \vs 30.3 for the class-specific counterpart on ResNet-50-C4. This further highlights the division of labor in our approach which largely decouples classification and segmentation.

\begin{table*}[t]
\tablestyle{3.5pt}{1.1}
\begin{tabular}{l|l|x{22}x{22}x{22}|x{22}x{22}x{22}}
 & backbone
 & AP$^\text{bb}$ & AP$^\text{bb}_{50}$ & AP$^\text{bb}_{75}$
 & AP$^\text{bb}_S$ & AP$^\text{bb}_M$ &  AP$^\text{bb}_L$\\ [.1em]
\shline
 Faster R-CNN+++ \cite{He2016} & ResNet-101-C4
  & 34.9 & 55.7 & 37.4 & 15.6 & 38.7 & 50.9\\
 Faster R-CNN w FPN \cite{Lin2017} & ResNet-101-FPN
  & 36.2 & 59.1 & 39.0 & 18.2 & 39.0 & 48.2\\
 Faster R-CNN by G-RMI \cite{Huang2017} & Inception-ResNet-v2 \cite{Szegedy2016a}
  & 34.7 & 55.5 & 36.7 & 13.5 & 38.1 & 52.0\\
 Faster R-CNN w TDM \cite{Shrivastava2016a} & Inception-ResNet-v2-TDM
  & 36.8 & 57.7 & 39.2 & 16.2 & 39.8 & \bd{52.1}\\
\hline
  Faster R-CNN, RoIAlign & ResNet-101-FPN
  & 37.3 & 59.6 & 40.3 & 19.8 & 40.2 & 48.8\\
 \bd{Mask R-CNN} & ResNet-101-FPN
  & 38.2 & 60.3 & 41.7 & 20.1 & 41.1 & 50.2\\
 \bd{Mask R-CNN} & ResNeXt-101-FPN
  & \bd{39.8} & \bd{62.3} & \bd{43.4} & \bd{22.1} & \bd{43.2} & {51.2}
\end{tabular}\vspace{1mm}
\caption{\textbf{Object detection} \emph{single-model} results (bounding box AP), \vs state-of-the-art on \texttt{test-dev}. Mask R-CNN using ResNet-101-FPN outperforms the base variants of all previous state-of-the-art models (the mask output is ignored in these experiments). The gains of Mask R-CNN over \cite{Lin2017} come from using RoIAlign (+1.1 AP$^\text{bb}$), multitask training (+0.9 AP$^\text{bb}$), and ResNeXt-101 (+1.6 AP$^\text{bb}$).}
\label{tab:final_bbox}\vspace{-4mm}
\end{table*}

\paragraph{RoIAlign:} An evaluation of our proposed \emph{RoIAlign} layer is shown in Table~\ref{tab:ablation:roialign}. For this experiment we use the ResNet-50-C4 backbone, which has stride 16. RoIAlign improves AP by about 3 points over RoIPool, with much of the gain coming at high IoU (AP$_{75}$). RoIAlign is insensitive to max/average pool; we use average in the rest of the paper.

Additionally, we compare with \emph{RoIWarp} proposed in MNC \cite{Dai2016} that also adopt bilinear sampling. As discussed in \S\ref{sec:maskrcnn}, RoIWarp still quantizes the RoI, losing alignment with the input. As can be seen in Table \ref{tab:ablation:roialign}, RoIWarp performs on par with RoIPool and much worse than RoIAlign. This highlights that proper alignment is key.

We also evaluate RoIAlign with a \emph{ResNet-50-C5} backbone, which has an even larger stride of 32 pixels. We use the same head as in Figure \ref{fig:head} (right), as the res5 head is not applicable. Table \ref{tab:ablation:roialign32} shows that RoIAlign improves mask AP by a massive 7.3 points, and mask AP$_{75}$ by 10.5 points (\emph{50\% relative improvement}). Moreover, we note that with RoIAlign, using \emph{stride-32} C5 features (30.9 AP) is more accurate than using stride-16 C4 features (30.3 AP, Table~\ref{tab:ablation:roialign}). RoIAlign largely resolves the long-standing challenge of using large-stride features for detection and segmentation.

Finally, RoIAlign shows a gain of 1.5 mask AP and 0.5 box AP when used with FPN, which has finer multi-level strides. For keypoint detection that requires finer alignment, RoIAlign shows large gains even with FPN (Table~\ref{tab:roialign_keypoint}).

\paragraph{Mask Branch:} Segmentation is a pixel-to-pixel task and we exploit the spatial layout of masks by using an FCN. In Table \ref{tab:ablation:maskhead}, we compare multi-layer perceptrons (MLP) and FCNs, using a ResNet-50-FPN backbone. Using FCNs gives a 2.1 mask AP gain over MLPs. We note that we choose this backbone so that the conv layers of the FCN head are not pre-trained, for a fair comparison with MLP.  

\subsection{Bounding Box Detection Results}

We compare Mask R-CNN to the state-of-the-art COCO \emph{bounding-box} object detection in Table~\ref{tab:final_bbox}. For this result, even though the full Mask R-CNN model is trained, only the classification and box outputs are used at inference (the mask output is ignored). Mask R-CNN using ResNet-101-FPN outperforms the base variants of all previous state-of-the-art models, including the single-model variant of G-RMI \cite{Huang2017}, the winner of the COCO 2016 Detection Challenge. Using ResNeXt-101-FPN, Mask R-CNN further improves results, with a margin of 3.0 points box AP over the best previous single model entry from \cite{Shrivastava2016a} (which used Inception-ResNet-v2-TDM).

As a further comparison, we trained a version of Mask R-CNN but \emph{without} the mask branch, denoted by ``Faster R-CNN, RoIAlign'' in Table \ref{tab:final_bbox}. This model performs better than the model presented in \cite{Lin2017} due to RoIAlign. On the other hand, it is 0.9 points box AP lower than Mask R-CNN. This gap of Mask R-CNN on box detection is therefore due solely to the benefits of multi-task training.

Lastly, we note that Mask R-CNN attains a small gap between its mask and box AP: \eg, 2.7 points between 37.1 (mask, Table~\ref{tab:final_mask}) and 39.8 (box, Table~\ref{tab:final_bbox}). This indicates that our approach largely closes the gap between object detection and the more challenging instance segmentation task.

\subsection{Timing}

\paragraph{Inference:} We train a ResNet-101-FPN model that shares features between the RPN and Mask R-CNN stages, following the 4-step training of Faster R-CNN \cite{Ren2015a}. This model runs at 195ms per image on an Nvidia Tesla M40 GPU (plus 15ms CPU time resizing the outputs to the original resolution), and achieves statistically the same mask AP as the unshared one. We also report that the ResNet-101-C4 variant takes $\app$400ms as it has a heavier box head (Figure \ref{fig:head}), so we do not recommend using the C4 variant in practice.

Although Mask R-CNN is fast, we note that our design is not optimized for speed, and better speed/accuracy trade-offs could be achieved \cite{Huang2017}, \eg, by varying image sizes and proposal numbers, which is beyond the scope of this paper.

\paragraph{Training:} Mask R-CNN is also fast to train. Training with ResNet-50-FPN on COCO \texttt{trainval35k} takes 32 hours in our synchronized 8-GPU implementation (0.72s per 16-image mini-batch), and 44 hours with ResNet-101-FPN. In fact, fast prototyping can be completed in \emph{less than one day} when training on the \texttt{train} set. We hope such rapid training will remove a major hurdle in this area and encourage more people to perform research on this challenging topic.

\begin{figure*}[t]
\centering
\includegraphics[width=1.0\linewidth]{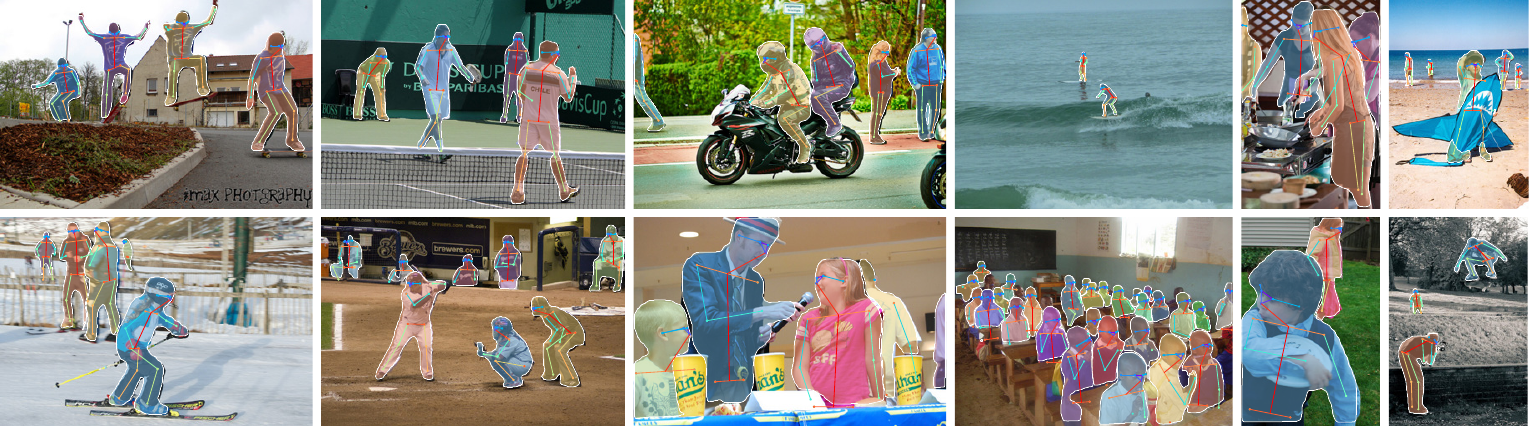}
\caption{Keypoint detection results on COCO \texttt{test} using Mask R-CNN (ResNet-50-FPN), with person segmentation masks predicted from the same model. This model has a keypoint AP of 63.1 and runs at 5 fps.}
\label{fig:results_keypoints}\vspace{-4mm}
\end{figure*}

\section{Mask R-CNN for Human Pose Estimation}\label{sec:keypoints}

Our framework can easily be extended to human pose estimation. We model a keypoint's location as a one-hot mask, and adopt Mask R-CNN to predict $K$ masks, one for each of $K$ keypoint types (\eg, left shoulder, right elbow). This task helps demonstrate the flexibility of Mask R-CNN.

We note that \emph{minimal} domain knowledge for human pose is exploited by our system, as the experiments are mainly to demonstrate the generality of the Mask R-CNN framework. We expect that domain knowledge (\eg, modeling structures \cite{Cao2017}) will be complementary to our simple approach.

\paragraph{Implementation Details:} We make minor modifications to the segmentation system when adapting it for keypoints. For each of the $K$ keypoints of an instance, the training target is a one-hot $m \x m$ binary mask where only a \emph{single} pixel is labeled as foreground. During training, for each visible ground-truth keypoint, we minimize the cross-entropy loss over an $m^2$-way softmax output (which encourages a single point to be detected). We note that as in instance segmentation, the $K$ keypoints are still treated independently.

We adopt the ResNet-FPN variant, and the keypoint head architecture is similar to that in Figure \ref{fig:head} (right). The keypoint head consists of a stack of eight 3$\x$3 512-d conv layers, followed by a deconv layer and 2$\x$ bilinear upscaling, producing an output resolution of 56$\x$56. We found that a relatively high resolution output (compared to masks) is required for keypoint-level localization accuracy.

Models are trained on all COCO \texttt{trainval35k} images that contain annotated keypoints. To reduce overfitting, as this training set is smaller, we train using image scales randomly sampled from [640, 800] pixels; inference is on a single scale of 800 pixels. We train for 90k iterations, starting from a learning rate of 0.02 and reducing it by 10 at 60k and 80k iterations. We use bounding-box NMS with a threshold of 0.5. Other details are identical as in \S\ref{sec:impl}.

\begin{table}[t]
\tablestyle{1.8pt}{1.2}
\begin{tabular}{l|x{22}x{22}x{22}|x{22}x{22}}
 & AP$^\text{kp}$ & AP$^\text{kp}_{50}$ & AP$^\text{kp}_{75}$
 & AP$^\text{kp}_M$ &  AP$^\text{kp}_L$\\ [.1em]
\shline
CMU-Pose+++ \cite{Cao2017} & 61.8 & 84.9 & 67.5 & 57.1 & 68.2 \\
G-RMI \cite{Papandreou2017}$^\dagger$ & 62.4 & 84.0 & 68.5 & \bd{59.1} & 68.1 \\
\hline
 \bd{Mask R-CNN}, \footnotesize keypoint-only & 62.7 & 87.0 & 68.4 & 57.4 & 71.1 \\
 \bd{Mask R-CNN}, \footnotesize keypoint \& mask & \bd{63.1} & \bd{87.3} & \bd{68.7} & {57.8} & \bd{71.4} \\
\end{tabular}\vspace{2mm}
\caption{\textbf{Keypoint detection} AP on COCO \texttt{test-dev}. Ours is a single model (ResNet-50-FPN) that runs at 5 fps. CMU-Pose+++ \cite{Cao2017} is the 2016 competition winner that uses multi-scale testing, post-processing with CPM \cite{Wei2016}, and filtering with an object detector, adding a cumulative $\app$5 points (clarified in personal communication). $^\dagger$: G-RMI was trained on COCO \emph{plus} MPII \cite{Andriluka2014} (25k images), using two models (Inception-ResNet-v2 for bounding box detection and ResNet-101 for keypoints).}
\label{tab:final_keypoint}\vspace{-2mm}
\end{table}

\paragraph{Main Results and Ablations:} We evaluate the person keypoint AP (AP$^\text{kp}$) and experiment with a ResNet-50-FPN backbone; more backbones will be studied in the appendix. Table~\ref{tab:final_keypoint} shows that our result (62.7 AP$^\text{kp}$) is 0.9 points higher than the COCO 2016 keypoint detection winner \cite{Cao2017} that uses a multi-stage processing pipeline (see caption of Table~\ref{tab:final_keypoint}). Our method is considerably simpler and faster.

\begin{table}[t]
\tablestyle{7pt}{1.1}
\begin{tabular}{l|x{22}x{22}x{22}}
 & AP$^\text{bb}_\text{\emph{person}}$ & AP$^\text{mask}_\text{\emph{person}}$
 & AP$^\text{kp}$ \\ [.1em]
\shline
Faster R-CNN & 52.5 & - & - \\
Mask R-CNN, mask-only & \bd{53.6} & \bd{45.8} & - \\
Mask R-CNN, keypoint-only & 50.7 & - & 64.2 \\
Mask R-CNN, keypoint \& mask & 52.0 & 45.1 & \bd{64.7} \\
\end{tabular}\vspace{2mm}
\caption{\textbf{Multi-task learning} of box, mask, and keypoint about the \emph{person} category, evaluated on \texttt{minival}. All entries are trained on the same data for fair comparisons. The backbone is ResNet-50-FPN. The entries with 64.2 and 64.7 AP on \texttt{minival} have \texttt{test-dev} AP of 62.7 and 63.1, respectively (see Table~\ref{tab:final_keypoint}).}
\label{tab:multitask_keypoint}\vspace{-2mm}
\end{table}

More importantly, we have \emph{a unified model that can simultaneously predict boxes, segments, and keypoints} while running at 5 fps. Adding a segment branch (for the person category) improves the AP$^\text{kp}$ to 63.1 (Table~\ref{tab:final_keypoint}) on \texttt{test-dev}. More ablations of multi-task learning on \texttt{minival} are in Table~\ref{tab:multitask_keypoint}. Adding the \emph{mask} branch to the box-only (\ie, Faster R-CNN) or keypoint-only versions consistently improves these tasks. However, adding the keypoint branch reduces the box/mask AP slightly, suggesting that while keypoint detection benefits from multitask training, it does not in turn help the other tasks. Nevertheless, learning all three tasks jointly enables a unified system to efficiently predict all outputs simultaneously (Figure~\ref{fig:results_keypoints}).

\begin{table}[t]
\tablestyle{2pt}{1.1}
\begin{tabular}{l|x{22}x{22}x{22}|x{22}x{22}}
 & AP$^\text{kp}$ & AP$^\text{kp}_{50}$ & AP$^\text{kp}_{75}$
 & AP$^\text{kp}_M$ &  AP$^\text{kp}_L$\\ [.1em]
\shline
 \emph{RoIPool} & 59.8 & 86.2 & 66.7 & 55.1 & 67.4 \\
 \emph{RoIAlign}~~~~ & \bd{64.2} & \bd{86.6} & \bd{69.7} & \bd{58.7} & \bd{73.0} \\
\end{tabular}\vspace{2mm}
\caption{\textbf{RoIAlign \vs RoIPool} for keypoint detection on \texttt{minival}. The backbone is ResNet-50-FPN.}
\label{tab:roialign_keypoint}\vspace{-2mm}
\end{table}

We also investigate the effect of \emph{RoIAlign} on keypoint detection (Table~\ref{tab:roialign_keypoint}).  Though this ResNet-50-FPN backbone has finer strides (\eg, 4 pixels on the finest level), RoIAlign still shows significant improvement over RoIPool and increases AP$^\text{kp}$ by 4.4 points. This is because keypoint detections are more sensitive to localization accuracy. This again indicates that alignment is essential for pixel-level localization, including masks and keypoints.

Given the effectiveness of Mask R-CNN for extracting object bounding boxes, masks, and keypoints, we expect it be an effective framework for other instance-level tasks.

\begin{table*}[t]
\tablestyle{4pt}{1.05}
\begin{tabular}{l|l|x{32}|x{22}x{22}|x{22}x{22}x{22}x{22}x{22}x{22}x{22}x{22}}
 & \multicolumn{1}{c|}{training data} & AP [\texttt{val}] & AP & AP$_{50}$
 & person & rider & car & truck & bus & train & mcycle & bicycle\\[.1em]
\shline
 InstanceCut \cite{Kirillov2017} & \texttt{fine} + \texttt{coarse}
  & 15.8 & 13.0 & 27.9 & 10.0 & 8.0 & 23.7 & 14.0 & 19.5 & 15.2 & 9.3 & 4.7 \\
 DWT \cite{Bai2017} & \texttt{fine}
  & 19.8 & 15.6 & 30.0 & 15.1 & 11.7 & 32.9 & 17.1 & 20.4 & 15.0 & 7.9 & 4.9 \\
 SAIS \cite{Hayder2017} & \texttt{fine}
  & - & 17.4 & 36.7 & 14.6 & 12.9 & 35.7 & 16.0 & 23.2 & 19.0 & 10.3 & 7.8 \\
 DIN \cite{Arnab2017} & \texttt{fine} + \texttt{coarse}
  & - & 20.0 & 38.8 & 16.5 & 16.7 & 25.7 & 20.6 & 30.0 & 23.4 & 17.1 & 10.1 \\
 SGN \cite{Liu2017} & \texttt{fine} + \texttt{coarse} & 29.2 & 25.0 & 44.9 & 21.8 &	20.1 &	39.4 &	24.8 &	33.2 &	30.8 &	17.7 &	12.4 \\
\hline
 Mask R-CNN & \texttt{fine}
  & 31.5 & 26.2 & 49.9 & 30.5 & 23.7 & 46.9 & 22.8 & 32.2 & 18.6 & 19.1 & 16.0 \\
 Mask R-CNN & \texttt{fine} + COCO
  & \bd{36.4} & \bd{32.0} & \bd{58.1} & \bd{34.8} & \bd{27.0} & \bd{49.1} & \bd{30.1} & \bd{40.9} & \bd{30.9} & \bd{24.1} & \bd{18.7} \\
\end{tabular}\vspace{1mm}
\caption{Results on Cityscapes \texttt{val} (`AP [\texttt{val}]' column) and \texttt{test} (remaining columns) sets. Our method uses ResNet-50-FPN.}\vspace{-3mm}
\label{tab:cityscapes}
\end{table*}

\appendix
\section*{Appendix A: Experiments on Cityscapes}

We further report instance segmentation results on the Cityscapes \cite{Cordts2016} dataset. This dataset has \texttt{fine} annotations for 2975 train, 500 val, and 1525 test images. It has 20k \texttt{coarse} training images without instance annotations, which we do \emph{not} use. All images are 2048$\times$1024 pixels. The instance segmentation task involves 8 object categories, whose numbers of instances on the \texttt{fine} training set are:

{\centering\resizebox{\columnwidth}{!}{
\tablestyle{.5pt}{1.1}\footnotesize
\begin{tabular}{x{28}|x{28}|x{28}|x{28}|x{28}|x{28}|x{28}|x{28}}
\multicolumn{8}{c}{}\\[-1em]
  person & rider & car & truck & bus & train & mcycle & bicycle\\
\hline
 17.9k & 1.8k & 26.9k & 0.5k & 0.4k & 0.2k & 0.7k & 3.7k
\end{tabular}}\vspace{.2em}}

\noindent Instance segmentation performance on this task is measured by the COCO-style mask AP (averaged over IoU thresholds); AP$_{50}$ (\ie, mask AP at an IoU of 0.5) is also reported.

\paragraph{Implementation:} We apply our Mask R-CNN models with the ResNet-FPN-50 backbone; we found the 101-layer counterpart performs similarly due to the small dataset size. We train with image scale (shorter side) randomly sampled from [800, 1024], which reduces overfitting; inference is on a single scale of 1024 pixels. We use a mini-batch size of 1 image per GPU (so 8 on 8 GPUs) and train the model for 24k iterations, starting from a learning rate of 0.01 and reducing it to 0.001 at 18k iterations. It takes $\app$4 hours of training on a single 8-GPU machine under this setting.

\paragraph{Results:} Table~\ref{tab:cityscapes} compares our results to the state of the art on the \texttt{val} and \texttt{test} sets. \emph{Without} using the \texttt{coarse} training set, our method achieves 26.2 AP on \texttt{test}, which is over 30\% relative improvement over the previous best entry (DIN \cite{Arnab2017}), and is also better than the concurrent work of SGN's 25.0 \cite{Liu2017}. Both DIN and SGN use \texttt{fine} + \texttt{coarse} data. Compared to the best entry using \texttt{fine} data only (17.4 AP), we achieve a $\app$50\% improvement.

\begin{figure}[t]
\centering
\includegraphics[width=1.0\linewidth]{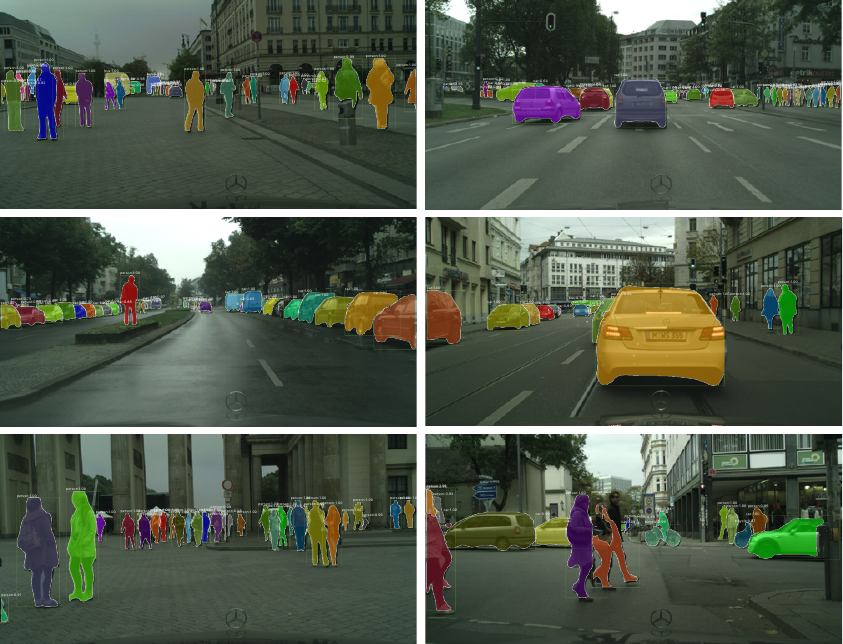}
\caption{Mask R-CNN results on Cityscapes \texttt{test} (32.0 AP). The bottom-right image shows a failure prediction.}
\label{fig:results_cityscapes}\vspace{-4mm}
\end{figure}

For the \emph{person} and \emph{car} categories, the Cityscapes dataset exhibits a large number of \emph{within}-category overlapping instances (on average 6 people and 9 cars per image). We argue that \emph{within-category overlap} is a core difficulty of instance segmentation. Our method shows massive improvement on these two categories over the other best entries (relative $\app$40\% improvement on \emph{person} from 21.8 to 30.5 and $\app$20\% improvement on \emph{car} from 39.4 to 46.9), even though our method does not exploit the \texttt{coarse} data.

A main challenge of the Cityscapes dataset is training models in a \emph{low-data} regime, particularly for the categories of \emph{truck}, \emph{bus}, and \emph{train}, which have about 200-500 training samples each. To partially remedy this issue, we further report a result using COCO pre-training. To do this, we initialize the corresponding 7 categories in Cityscapes from a pre-trained COCO Mask R-CNN model (\emph{rider} being randomly initialized). We fine-tune this model for 4k iterations in which the learning rate is reduced at 3k iterations, which takes $\app$1 hour for training given the COCO model.

The COCO pre-trained Mask R-CNN model achieves 32.0 AP on \texttt{test}, almost a 6 point improvement over the \texttt{fine}-only counterpart. This indicates the important role the amount of training data plays. It also suggests that methods on Cityscapes might be influenced by their \emph{low-shot} learning performance. We show that using COCO pre-training is an effective strategy on this dataset.

Finally, we observed a bias between the \texttt{val} and \texttt{test} AP, as is also observed from the results of \cite{Kirillov2017,Bai2017,Liu2017}. We found that this bias is mainly caused by the \emph{truck}, \emph{bus}, and \emph{train} categories, with the \texttt{fine}-only model having \texttt{val}/\texttt{test} AP of 28.8/22.8, 53.5/32.2, and 33.0/18.6, respectively. This suggests that there is a \emph{domain shift} on these categories, which also have little training data. COCO pre-training helps to improve results the most on these categories; however, the domain shift persists with 38.0/30.1, 57.5/40.9, and 41.2/30.9 \texttt{val}/\texttt{test} AP, respectively. Note that for the \emph{person} and \emph{car} categories we do not see any such bias (\texttt{val}/\texttt{test} AP are within $\pm1$ point).

Example results on Cityscapes are shown in Figure~\ref{fig:results_cityscapes}.

\section*{Appendix B: Enhanced Results on COCO}

As a general framework, Mask R-CNN is compatible with complementary techniques developed for detection/segmentation, including improvements made to Fast/Faster R-CNN and FCNs. In this appendix we describe some techniques that improve over our original results. Thanks to its generality and flexibility, Mask R-CNN was used as the framework by the three winning teams in the COCO 2017 instance segmentation competition, which all significantly outperformed the previous state of the art.

\begin{table}[t]
\resizebox{\columnwidth}{!}{\tablestyle{2pt}{1.05}
\begin{tabular}{l|l|x{22}x{22}x{22}|x{22}x{22}x{22}}
description & backbone & AP & AP$_{50}$ & AP$_{75}$ & AP$^\text{bb}$ & AP$^\text{bb}_{50}$ & AP$^\text{bb}_{75}$ \\ [.1em]
\shline
original baseline & X-101-FPN & 36.7 & 59.5 & 38.9 & 39.6 & 61.5 & 43.2 \\
$+$ updated baseline & X-101-FPN & 37.0 & 59.7 & 39.0 & 40.5 & 63.0 & 43.7 \\
$+$ e2e training & X-101-FPN & 37.6 & 60.4 & 39.9 & 41.7 & 64.1 & 45.2 \\
$+$ ImageNet-5k & X-101-FPN & 38.6 & 61.7 & 40.9 & 42.7 & 65.1 & 46.6 \\
$+$ train-time augm. & X-101-FPN & 39.2 & 62.5 & 41.6 & 43.5 & 65.9 & 47.2 \\
$+$ deeper & X-152-FPN & 39.7 & 63.2 & 42.2 & 44.1 & 66.4 & 48.4 \\
$+$ Non-local \cite{Wang2017} & X-152-FPN-NL & \bd{40.3} & \bd{64.4} & \bd{42.8} & \bd{45.0} & \bd{67.8} & \bd{48.9} \\\hline
$+$ test-time augm. & X-152-FPN-NL & \bd{41.8} & \bd{66.0} & \bd{44.8} & \bd{47.3} & \bd{69.3} & \bd{51.5} \\
\end{tabular}}\vspace{1mm}
\caption{\textbf{Enhanced detection results} of Mask R-CNN on COCO \texttt{minival}. Each row adds an extra component to the above row. We denote ResNeXt model by `X' for notational brevity.}
\label{tab:enhanced}\vspace{-3mm}
\end{table}

\subsection*{Instance Segmentation and Object Detection}

We report some enhanced results of Mask R-CNN in Table~\ref{tab:enhanced}. Overall, the improvements increase mask AP 5.1 points (from 36.7 to 41.8) and box AP 7.7 points (from 39.6 to 47.3). Each model improvement increases both mask AP and box AP consistently, showing good generalization of the Mask R-CNN framework. We detail the improvements next. These results, along with future updates, can be reproduced by our released code at \url{https://github.com/facebookresearch/Detectron}, and can serve as higher baselines for future research.

\textit{Updated baseline:} We start with an updated baseline with a different set of hyper-parameters. We lengthen the training to 180k iterations, in which the learning rate is reduced by 10 at 120k and 160k iterations. We also change the NMS threshold to 0.5 (from a default value of 0.3). The updated baseline has 37.0 mask AP and 40.5 box AP.

\textit{End-to-end training:} All previous results used stage-wise training, \ie, training RPN as the first stage and Mask R-CNN as the second. Following \cite{Ren2017}, we evaluate end-to-end (`e2e') training that jointly trains RPN and Mask R-CNN. We adopt the `approximate' version in \cite{Ren2017} that only computes partial gradients in the RoIAlign layer by ignoring the gradient \wrt RoI coordinates. Table~\ref{tab:enhanced} shows that e2e training improves mask AP by 0.6 and box AP by 1.2.

\textit{ImageNet-5k pre-training:} Following \cite{Xie2017}, we experiment with models pre-trained on a 5k-class subset of ImageNet (in contrast to the standard 1k-class subset). This 5$\times$ increase in pre-training data improves both mask and box 1 AP. As a reference, \cite{Sun2017} used $\app$250$\times$ more images (300M) and reported a 2-3 box AP improvement on their baselines.

\textit{Train-time augmentation:} Scale augmentation at train time further improves results. During training, we randomly sample a scale from [640, 800] pixels and we increase the number of iterations to 260k (with the learning rate reduced by 10 at 200k and 240k iterations). Train-time augmentation improves mask AP by 0.6 and box AP by 0.8.

\textit{Model architecture:} By upgrading the 101-layer ResNeXt to its 152-layer counterpart \cite{He2016}, we observe an increase of 0.5 mask AP and 0.6 box AP. This shows a deeper model can still improve results on COCO.

Using the recently proposed \emph{non-local} (NL) model \cite{Wang2017}, we achieve 40.3 mask AP and 45.0 box AP. This result is without test-time augmentation, and the method runs at 3fps on an Nvidia Tesla P100 GPU at test time.

\textit{Test-time augmentation:} We combine the model results evaluated using scales of [400, 1200] pixels with a step of 100 and on their horizontal flips. This gives us a single-model result of 41.8 mask AP and 47.3 box AP.

The above result is the foundation of our submission to the COCO 2017 competition (which also used an ensemble, not discussed here). The first three winning teams for the instance segmentation task were all reportedly based on an extension of the Mask R-CNN framework.

\begin{table}[t]
\resizebox{\columnwidth}{!}{\tablestyle{3pt}{1.05}
\begin{tabular}{l|l|x{22}x{22}x{22}|x{22}x{22}}
description & backbone & AP$^\text{kp}$ & AP$^\text{kp}_{50}$ & AP$^\text{kp}_{75}$
 & AP$^\text{kp}_M$ & AP$^\text{kp}_L$\\ [.1em]
\shline
original baseline & R-50-FPN & 64.2 & 86.6 & 69.7 & 58.7 & 73.0 \\
$+$ updated baseline & R-50-FPN & 65.1 & 86.6 & 70.9 & 59.9 & 73.6 \\
$+$ deeper & R-101-FPN & 66.1 & 87.7 & 71.7 & 60.5 & 75.0\\
$+$ ResNeXt & X-101-FPN & 67.3 & 88.0 & 73.3 & 62.2 & 75.6 \\
$+$ data distillation \cite{Radosavovic2017} & X-101-FPN & \bd{69.1} & \bd{88.9} & \bd{75.3} & \bd{64.1} & \bd{77.1} \\
\hline
$+$ test-time augm. & X-101-FPN & \bd{70.4} & \bd{89.3} & \bd{76.8} & \bd{65.8} & \bd{78.1} \\
\end{tabular}}\vspace{1mm}
\caption{\textbf{Enhanced keypoint results} of Mask R-CNN on COCO \texttt{minival}. Each row adds an extra component to the above row. Here we use only keypoint annotations but no mask annotations. We denote ResNet by `R' and ResNeXt by `X' for brevity.}
\label{tab:enhanced_keypoint}\vspace{-3mm}
\end{table}

\subsection*{Keypoint Detection}

We report enhanced results of keypoint detection in Table~\ref{tab:enhanced_keypoint}. As an updated baseline, we extend the training schedule to 130k iterations in which the learning rate is reduced by 10 at 100k and 120k iterations. This improves AP$^\text{kp}$ by about 1 point. Replacing ResNet-50 with ResNet-101 and ResNeXt-101 increases AP$^\text{kp}$ to 66.1 and 67.3, respectively.

With a recent method called \emph{data distillation} \cite{Radosavovic2017}, we are able to exploit the additional 120k \emph{unlabeled} images provided by COCO. In brief, data distillation is a self-training strategy that uses a model trained on labeled data to predict annotations on unlabeled images, and in turn updates the model with these new annotations. Mask R-CNN provides an effective framework for such a self-training strategy. With data distillation, Mask R-CNN AP$^\text{kp}$ improve by 1.8 points to 69.1. We observe that Mask R-CNN can benefit from extra data, even if that data is \emph{unlabeled}.

By using the same test-time augmentation as used for instance segmentation, we further boost AP$^\text{kp}$ to 70.4.

\paragraph{Acknowledgements:} We would like to acknowledge Ilija Radosavovic for contributions to code release and enhanced results, and the Caffe2 team for engineering support.

{\small\bibliographystyle{ieee}\bibliography{maskrcnn_arxiv.bib}}

\end{document}